\documentclass[preprint]{elsarticle}

\usepackage{graphicx}%
\usepackage{subfig}
\usepackage{url}
\usepackage{lineno}

\journal{Renewable Energy}
\begin{document}
\begin{frontmatter}

\title{Short-Term Predictability of Photovoltaic Production over Italy} 
        
\author[ENEA]{Matteo De Felice}
\author[ENEA,EURAC]{Marcello Petitta}
\author[ENEA]{Paolo M. Ruti}

\address[ENEA]{Casaccia R.C., ENEA 
Energy and Environment Modelling Technical Unit, Rome,
Italy e-mail: \{matteo.defelice\}@enea.it
}
\address[EURAC]{Institute for Applied Remote Sensing, EURAC, Viale  Druso 1, Bolzano/Bozen Italy %
}

\begin{abstract}
Photovoltaic (PV) power production increased drastically in Europe throughout the last years. About the 6\% of electricity in Italy comes from PV and for an efficient management of the power grid an accurate and reliable forecasting of production would be needed. Starting from a dataset of electricity production of 65 Italian solar plants for the years 2011-2012 we investigate the possibility to forecast daily production from one to ten days of lead time without using on site measurements.
Our study is divided in two parts: an assessment of the predictability of meteorological variables using weather forecasts and an analysis on the application of data-driven modelling in predicting solar power production. We calibrate a SVM model using available observations and then we force the same model with the predicted variables from weather forecasts with a lead time from one to ten days.
As expected, solar power production is strongly influenced by cloudiness and clear sky, in fact we observe that while during summer we obtain a general error under the 10\% (slightly lower in south Italy), during winter the error is abundantly above the 20\%.

\end{abstract}

\end{frontmatter}
\section{Introduction}

Europe is experiencing a growing penetration of photovoltaic (PV) production, in particular Italy that in 2012 had almost $480\,000$ PV plants (16.4 GW of total installed power) \cite{gse2012}, 44\% more than 2011. Modelling of daily electricity generation of a PV power system can be useful for an effective management and balancing of a power grid, supporting real-time operations especially in countries with a lot of solar energy potential. Forecasting the expected PV power production could in fact help to deal with its intermittency, mainly due to weather conditions. Moreover, short-term forecasting information can also be valuable for electric market operators. 


Production of a PV plant can be modelled in two ways: with a mathematical model and a data-driven approach, the latter often called black-box modelling. Both the approaches have their pros and cons, the former can be more accurate but in addition to weather variables (incoming solar radiation, air temperature, wind speed, etc.) it needs solar panel characteristics (technology, area, orientation, etc.). The black-box approach does not require information about the typology of PV panel but it needs long time-series of input and output variables to calibrate a reliable model. In our work, we use a Support Vector Machine (SVM, briefly introduced later in Sec. \ref{sec:modeling}) to perform the prediction of daily production using both solar radiation and temperature information. The choice of a black-box approach is due to the absence of both detailed information about solar panel characteristics and on-site measurements of solar irradiance and air temperature. 

SVMs have been already used for similar applications, Zeng \& Qiao \cite{zeng13} tested a SVM-based approach using data from three different sites outperforming both autoregressive and neural network-based models; Bouzerdoum et al. \cite{bouzerdoum13} proposed a hybrid SARIMA-SVM approach which performed better than both the single models in predicting hourly power output of a small PV plant. More in general, black-box methods are common for forecasting applications related to solar power and solar radiation (e.g., see Pedro \& Coimbra \cite{pedro12}). 

Our work is based on daily power production data of 65 grid-connected PV systems on Italy during the period 2011-2012. For each plant a SVM model has been  built and tested with the best available weather observations of solar radiation and air temperature, respectively provided by CM-SAF satellite and weather stations. Then, the same SVM models are used for forecasting power production using as inputs data the weather forecasts of solar radiation and temperature. 

In the next section, we introduce and describe weather and production data for modelling and forecasting parts, respectively presented in Section \ref{sec:modeling} and \ref{sec:forecast}. For a better comprehension of the forecasting results, we also analyse the predictability of solar radiation and temperature provided by weather forecasts in Sections \ref{sec:solarpred} and \ref{sec:temppred}. The final section provides a summary and conclusion.

\section{Data}
In this work a data-driven approach has been chosen, mainly due to the unavailability of detailed data about power plants and weather measurements. The effectiveness of a data-driven approach, as the name suggests, strongly relies in the appropriateness and quality of input/output data. Input data are here weather variables, solar radiation and air temperature, while the output variable is the electricity production. 
Solar radiation is converted into electricity by photovoltaic modules and for this reason the choice of surface incoming solar radiation as model input is obvious. Air temperature is also an important variable:  solar panels efficiency is sensitive to module temperature, depending of the specific equipment when it exceeds a threshold (generally about $25 ^{\circ}$C) the panel efficiency begins to drop. For an improved modelling of the module temperature  the cooling effect of the wind also should be taken into account (as described in Schwingshackl et al. \cite{schwingshackl13}) and its inclusion is left for future work.

\subsection{Meteorological Data}
\label{ssec:weather}

\begin{figure}
\centering
\subfloat[Average] {
\label{fig:avgssr}
	\includegraphics[width=2.5in]{{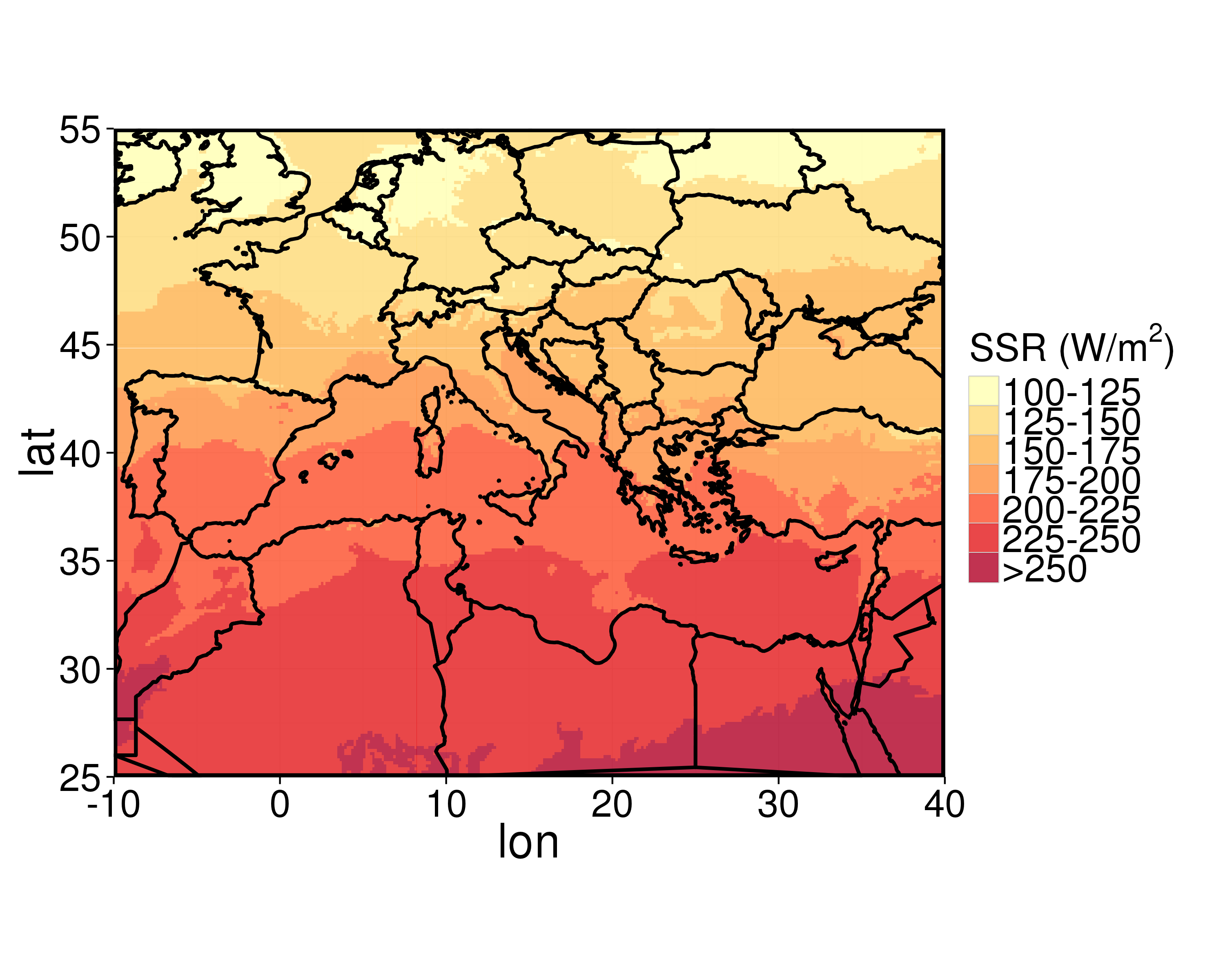}}
}
\subfloat[Coefficient of Variation] {
\label{fig:cvssr}
	\includegraphics[width=2.5in]{{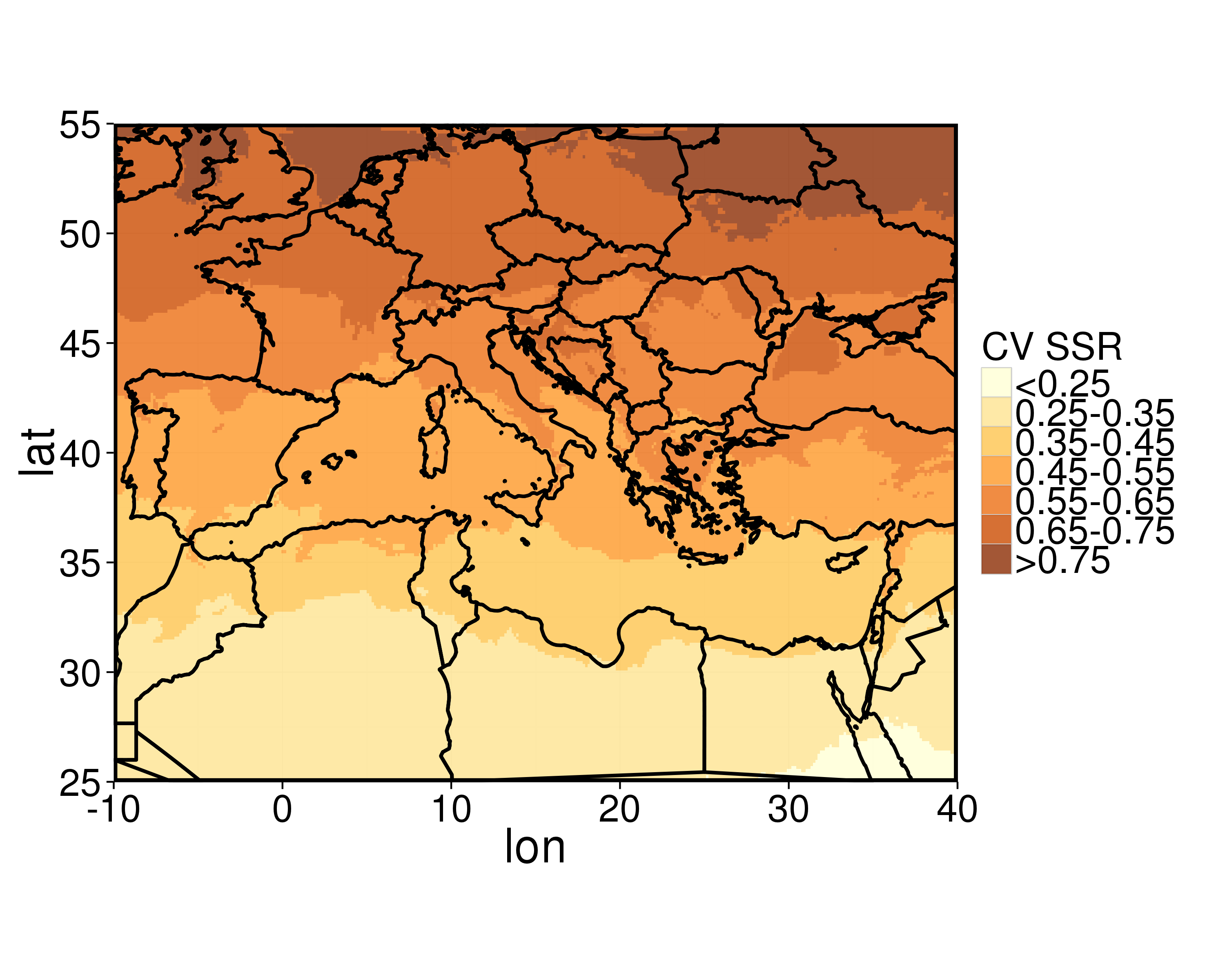}}
}
\caption{Surface Solar Radiation statistics for the years 2011-2012 from CM-SAF Satellite observations. Coefficient of variation measures the variability of the solar	radiation, we can observe as the Northern Europe show a higher variability (generally a $\mathrm{CV} > 0.55$ above the $45^{\circ}$ of latitude) and lower average solar radiation than the Southern part of the continent. }

\end{figure}

Solar radiation measurements used in this paper are obtained from the Satellite Application Facility on Climate Monitoring (CM-SAF) \cite{schulz09}, part of EUMETSAT's SAF Network. Considered variable is the surface incoming shortwave (SIS) radiation on the Meteosat (MSG) full disk. In Figure \ref{fig:avgssr} is visible the average daily solar radiation and its coefficient of variation (Figure \ref{fig:cvssr}), i.e. the ratio between standard deviation and average.

For the air temperature, we instead consider the E-OBS gridded dataset \cite{haylock08}, a land-only high-resolution temperature dataset obtained interpolating on a $0.25 ^{\circ}$ regular grid the available meteorological stations (4200 stations at the latest release made available in October 2013). 

Weather forecast of solar radiation and temperature data are provided by the ECMWF Integrated Forecasting System (IFS) which runs twice per day with a resolution of 16 km. 

\begin{table}[h]
\begin{tabular}{|l|l|l|}
\hline
 & Observed & Forecast  \\ \hline
2-m temperature &  E-OBS ($\sim$ 25 km) & ECMWF IFS ($\sim$ 16 km)\\
Downward solar radiation & CM-SAF ($\sim$ 5 km) & ECMWF IFS ($\sim$ 16 km)\\ \hline
\end{tabular}

\caption{Summary of weather datasets used in this work}
\label{tab:datasets}
\end{table}

In Table \ref{tab:datasets} are summarised all the data sources used in this paper.  


\subsection{Production Data}
\label{ssec:proddata}
In this work we consider 65 different PV power plants located in different Italian regions. We divided the plants in two groups: North and South. In the first group (North) we have all the PV plants above the $44^{\circ} \, 50'$ latitude, 34 PV plants with a total of 127 MW of installed capacity. Remaining plants are in the other group (South), 31 PV plants with a total of 288 MW. 

For each plant we have a time-series of daily power production of variable length, between 18 and 24 months (550--731 daily samples). 

\section{Daily Predictability of Meteorological Data}
\label{sec:meteopred}

In this section we analyze the capability of the ECMWF numerical weather prediction model to forecasts the two main predictors for solar power production: solar radiation and air temperature. Both the meteorological variables are provided by the ECMWF global forecast model, which data is available on $0.25^{\circ}$ grid and with a time step of 3 hours up to ten days in advance. 

An assessment on the forecasting skills of ECMWF model can be found in Richardson et al. \cite{richardson13}. Other studies on the use of solar radiation forecasts can be found in Lorenz et al. \cite{lorenz09} and Mathiesen \& Kleissl \cite{mathiesen11}.

\subsection{Solar Radiation}
\label{sec:solarpred}
ECMWF operational deterministic forecasts are issued every day and they provide hourly estimation of several variables up to ten days. We used the surface solar radiation downwards variable, i.e. the incident shortwave radiation, accumulated over the day. 

We compared the forecasts with the values measured by CM-SAF satellite data for the years 2011-2012. 

\begin{figure}
\centering
\includegraphics[width=3in]{{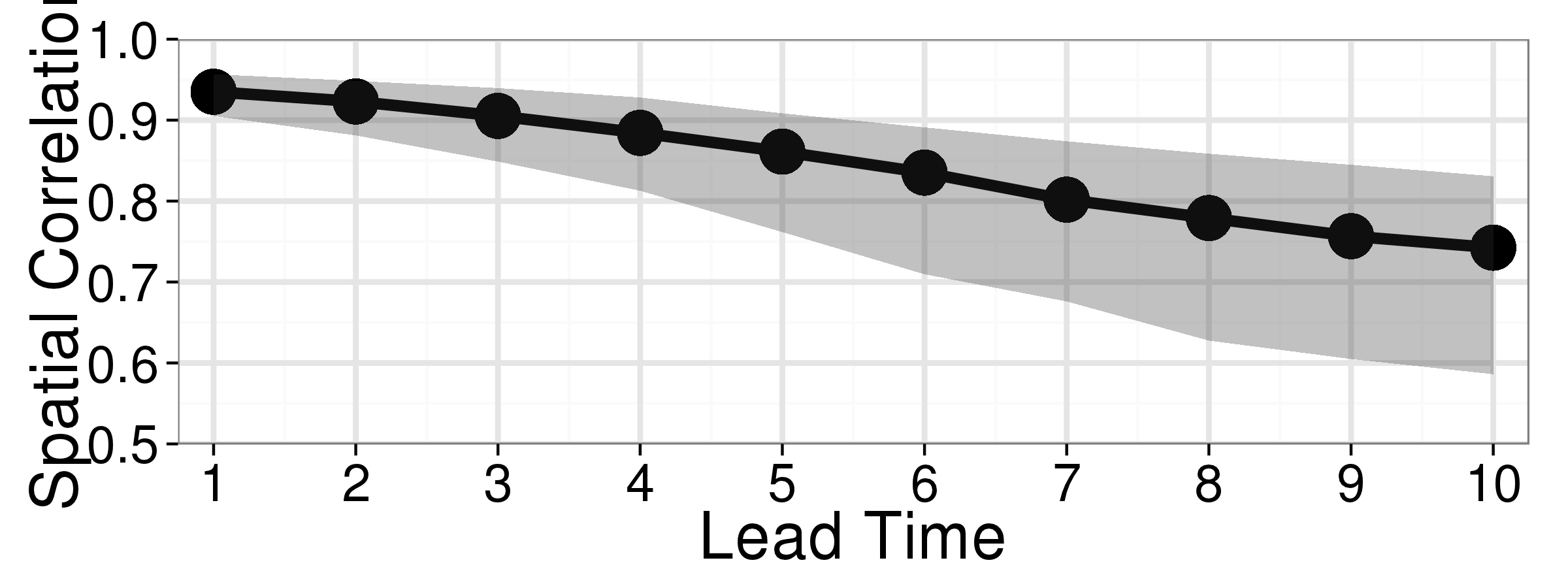}}
\caption{Average spatial correlation for the period 2011-2012 on the entire domain between operational forecasts and satellite measurements of solar radiation. Shaded area represents the interquartile range (IQR) for each lead time. We observe an average decrease of correlation of 2.5\% and an increment of IQR of 20\% for each lead-time. }
\label{fig:spatcorr}
\end{figure}

In Figure \ref{fig:spatcorr} we can observe the spatial correlation between forecasts and satellite data on the entire domain as a function of the lead time of forecast. 

\begin{figure}
\centering
\subfloat[Actual SSR - satellite] {
	\includegraphics[width=2.5in]{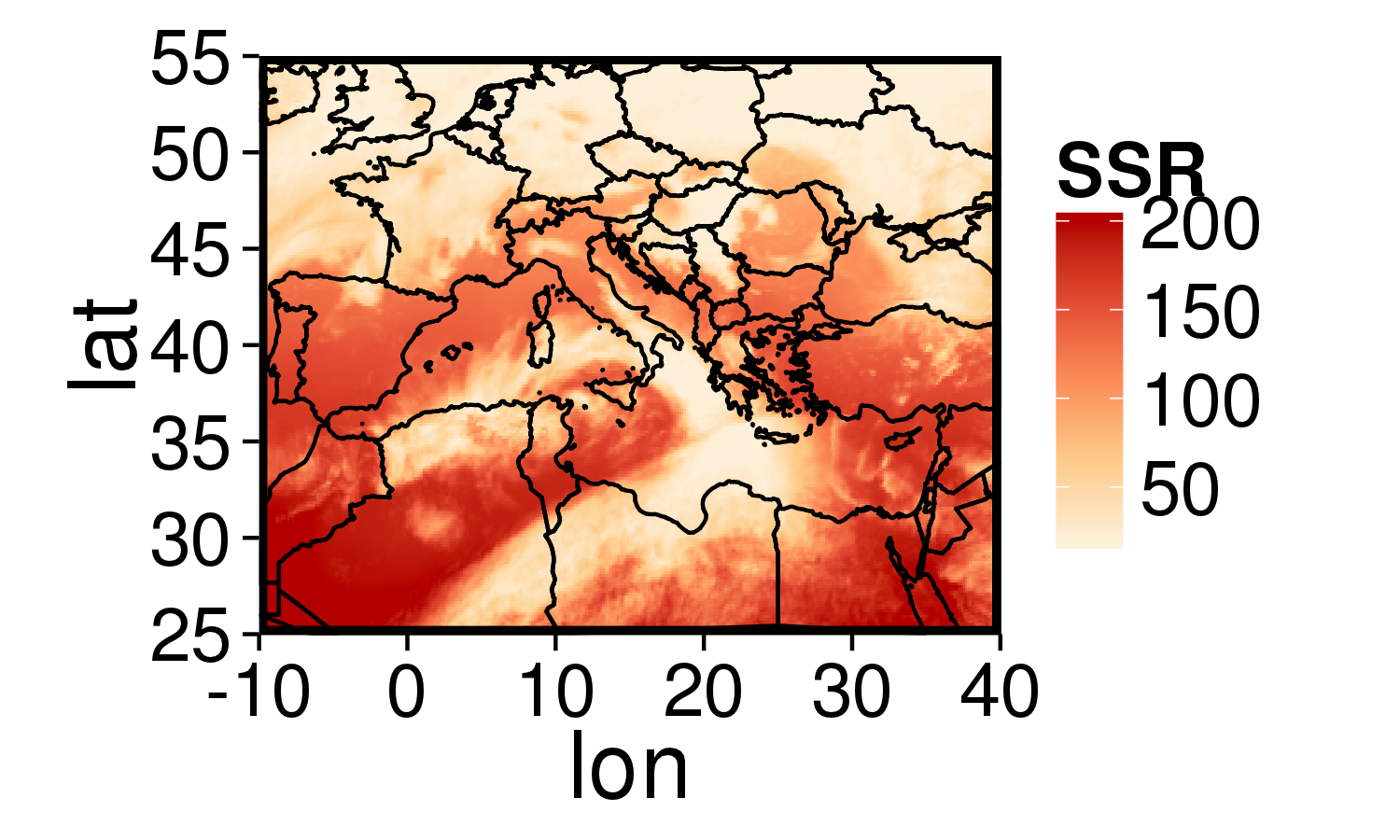}
} 
\subfloat[One day ahead] {
	\label{sfig:day01}
	\includegraphics[width=2.5in]{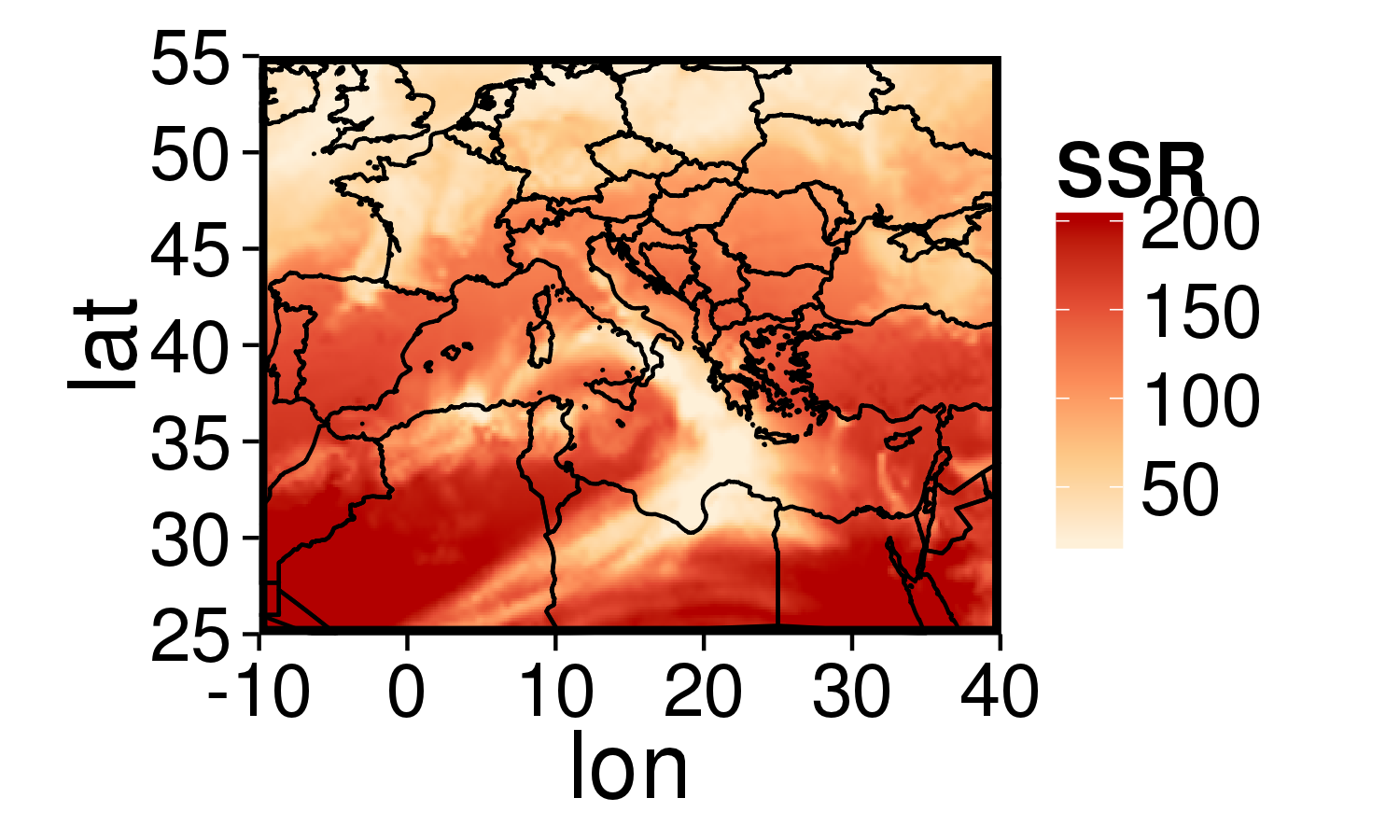}

} \hfill
\subfloat[Five days ahead] {
	\label{sfig:day05}
	\includegraphics[width=2.5in]{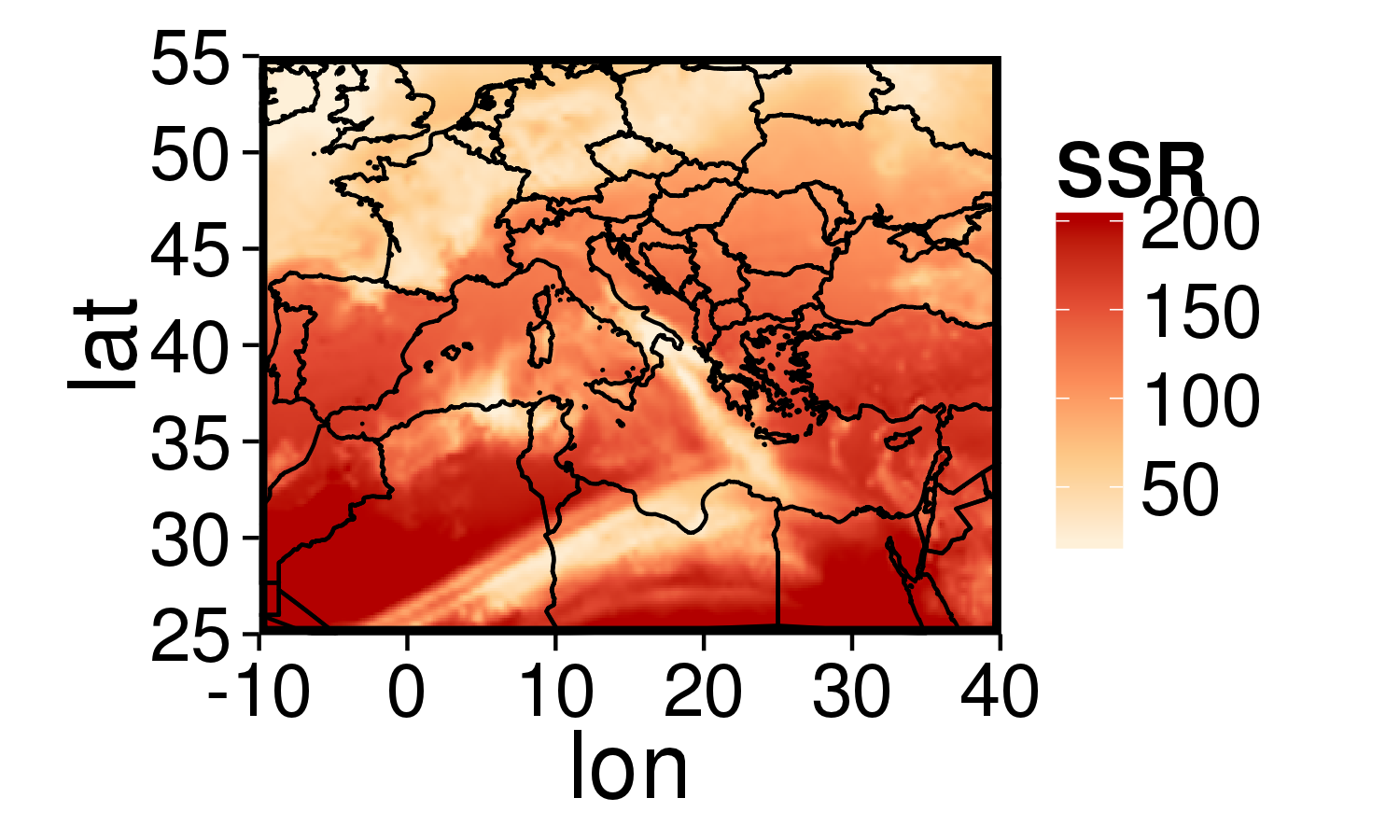}	
}
\subfloat[Ten days ahead] {
	\label{sfig:day10}
	\includegraphics[width=2.5in]{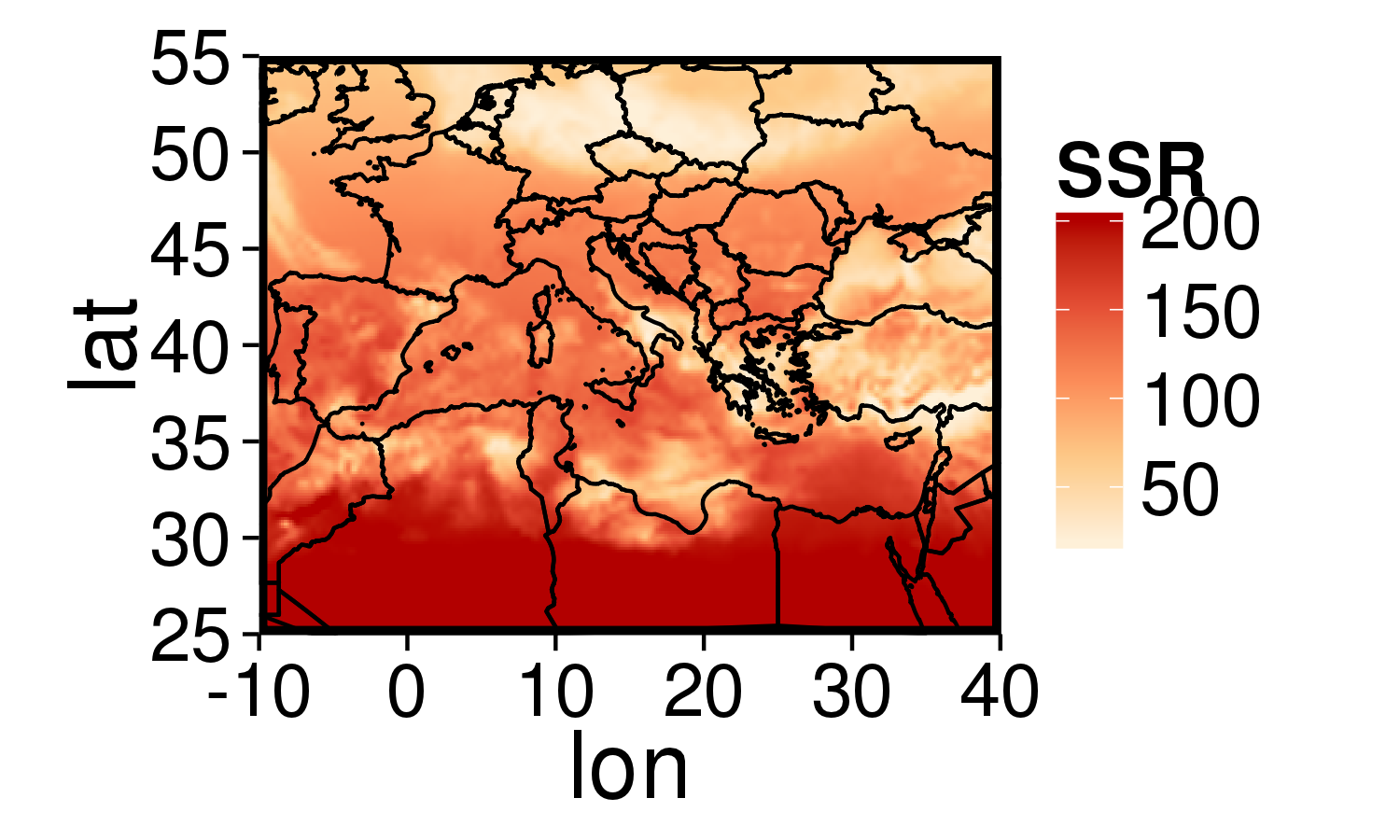}
} \hfill

\caption{Example for a specific day (2/2/2011) of solar radiation forecasts provided by ECMWF operational forecasts with one, five and ten days of lead-time. The spatial correlations of the shown forecasts with the observations are respectively 0.93, 0.90 and 0.67. }
\label{fig:exforecast}
\end{figure}

To better give an idea of the forecast quality, in Figure \ref{fig:exforecast} we show an example of a specific day forecast with three different lead times: one day (Fig. \ref{sfig:day01}, correlation of 0.93), five days (Fig. \ref{sfig:day05}, correlation 0.90) and ten days (Fig. \ref{sfig:day10}, correlation 0.67). 

\begin{figure}
\centering
\includegraphics[width=4.5in]{{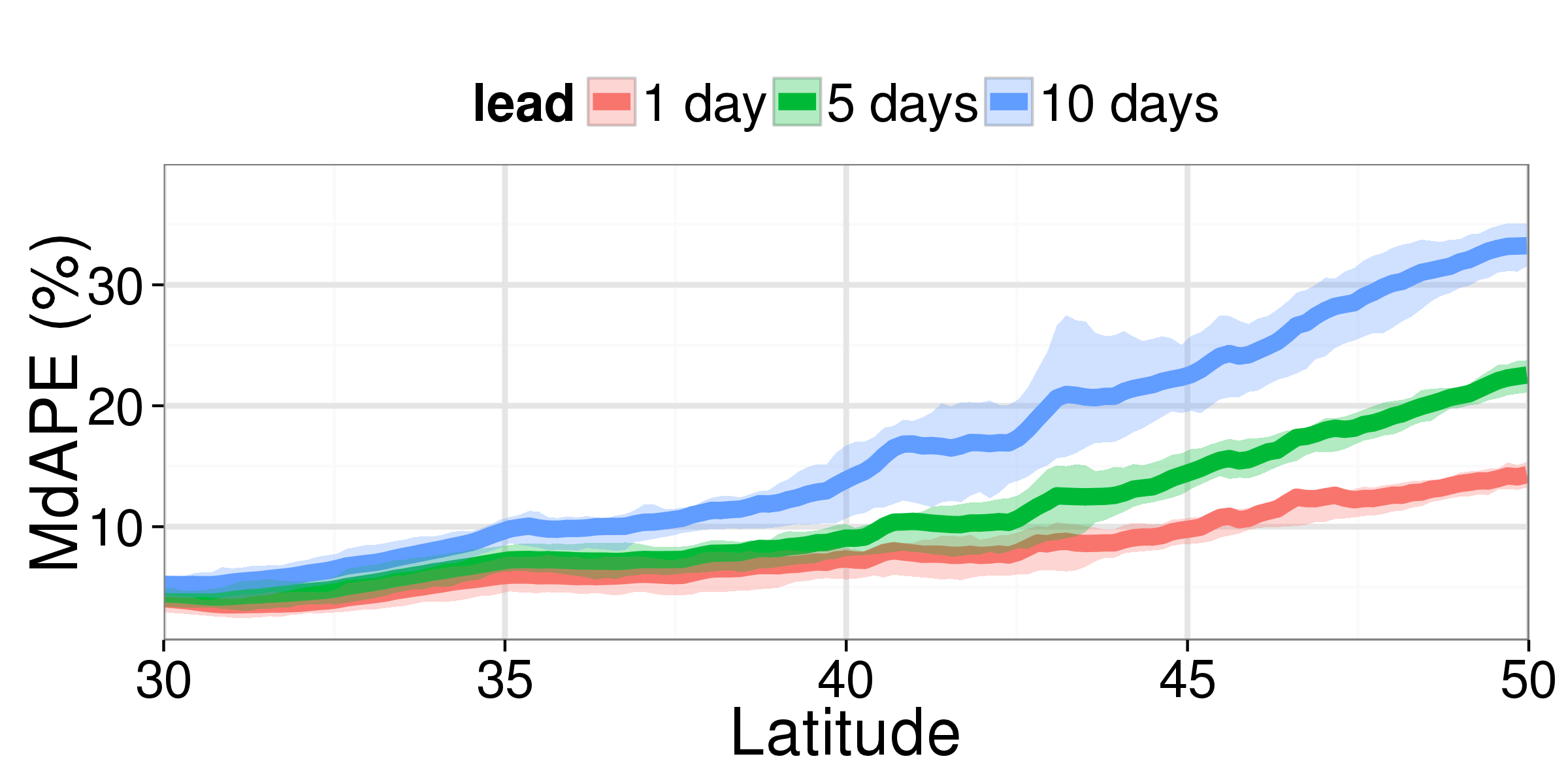}}
\caption{Error on solar radiation forecast versus latitude over Europe with selected lead times (one, five, and ten days). Shaded area represents the interquartile range (IQR). The range 30$^\circ$-35$^\circ$ is related to the North Africa and East Mediterranean where the solar radiation variability is low, in this case in fact the errors for the three lead-times are close to each other. Instead the range 40$^\circ$ and 45$^\circ$ includes the majority of the European mountain areas (Alps, Pyrenees, Carpathians, Balkans), in fact we observe an large forecast error variability (i.e. high IQR).}
\label{fig:latvsmdape}
\end{figure}

Solar radiation exhibits a clear seasonal cycle and for this reason absolute error measures (e.g. RMSE) might be misleading. We decide to use a percentage error measures, the Median Absolute Percentage Error (MdAPE) defined as:
\begin{equation}
 \mathrm{MdAPE} = \mathrm{median}(| 100 (\hat{y}_t - y_t)/y_t|)
 \end{equation} 
where $y_t$ is the observed value and $\hat{y}_t$ the estimation at time $t$.

Figure \ref{fig:latvsmdape} illustrates the MdAPE of the predicted solar radiation with respect to the latitude for three lead times (1, 5, 10 days, the other lead times have been omitted for sake of clarity). It is evident how the prediction error is related to the lead time, with one day the average MdAPE on the entire domain (30-50 $^{\circ}$ latitude) is $8.25\%$, with five days is $11.59\%$ and at ten days is $17.04\%$. 

We can observe how the performance of the forecast decreases at the high latitudes, due to the higher weather variability as also shown in Fig. \ref{fig:cvssr}. 

\begin{figure}
\centering
\subfloat[NORTH ITALY] {
	\label{sfig:ssrdensitynorth}
	\includegraphics[height=1.25in]{{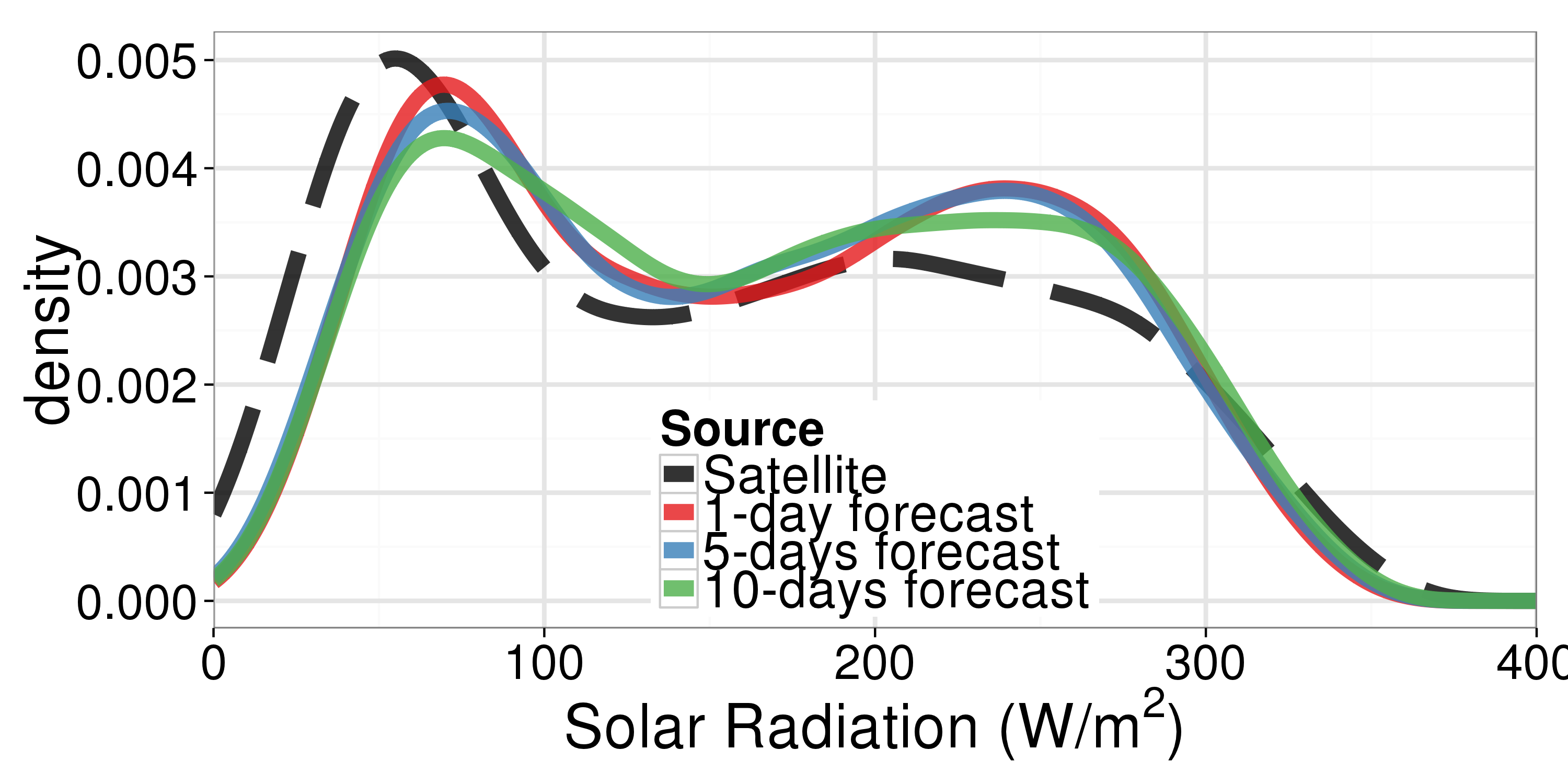}}
} \hfill
\subfloat[SOUTH ITALY] {
	\label{sfig:ssrdensitysouth}
	\includegraphics[height=1.25in]{{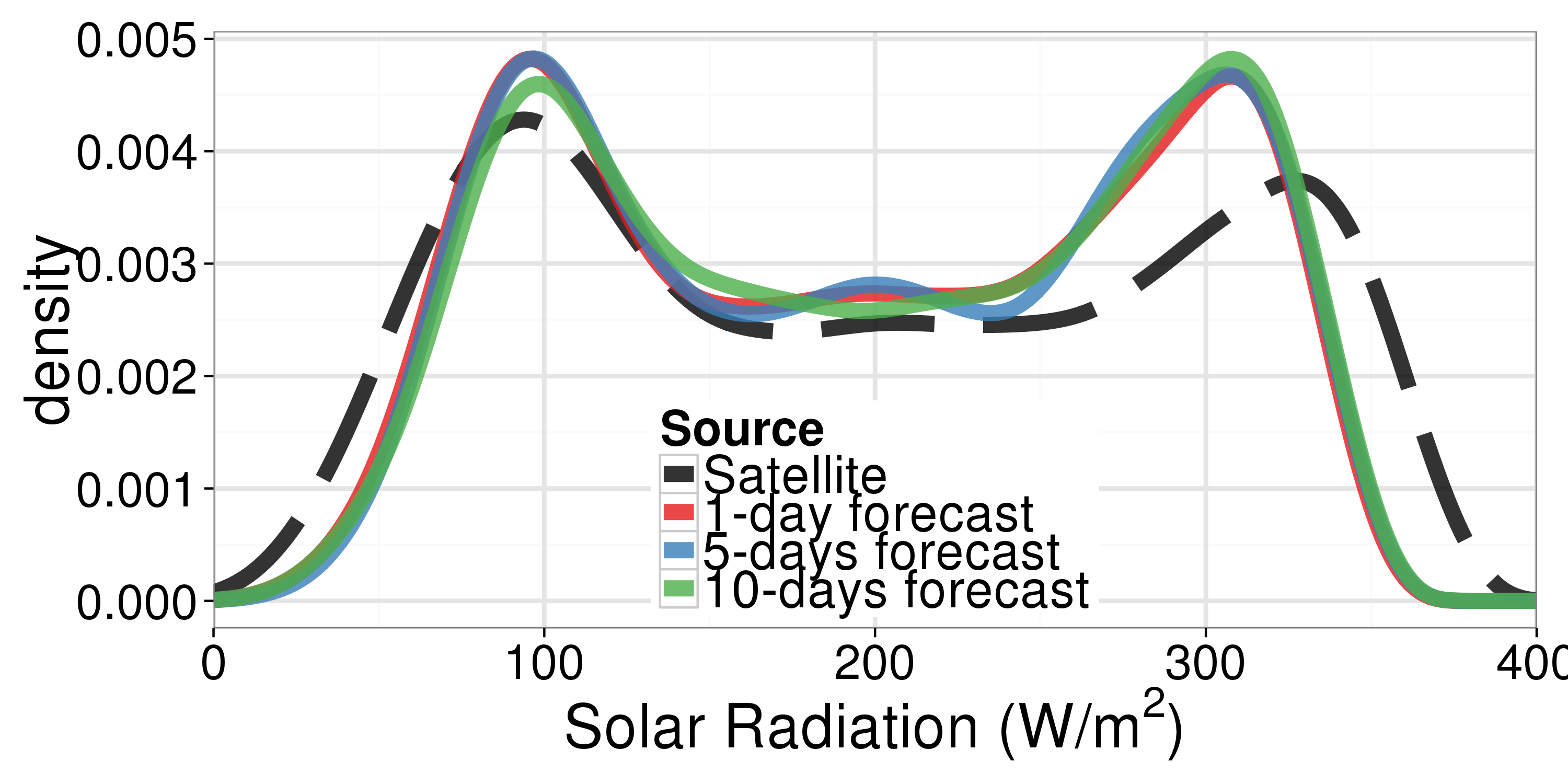}}
}
\caption{Comparison of Gaussian kernel density estimation of the observed solar radiation with the predictions at three lead-times (one, five and ten days). We can see that the weather forecasts tend to overestimate the ``winter'' (left one) peak in the North of Italy and to underestimate the ``summer'' peak (right one) in the South part.}
\label{fig:densitySSR}
\end{figure}

According to the North/South classification proposed in Section \ref{ssec:proddata}, in Figure \ref{fig:densitySSR} we show the density plot of solar radiation provided by CM-SAF (satellite) and by the forecast at one, five and ten days of lead time. Looking at the density plot for the North Italy (Fig. \ref{sfig:ssrdensitynorth}), we can quickly see the difference among the three lead times in describing the two peaks, especially for the minor one. Observing the density comparison for the South Italy (Fig. \ref{sfig:ssrdensitysouth}) we instead see how the three lead times show a similar distribution. It can be seen also that for the South Italy the forecasts tend to underestimate the highest peak. 

\subsection{Air Temperature}
\label{sec:temppred}
As for the downwards solar radiation, we analyze the predictability of air temperature provided by ECWMF deterministic forecasts by comparing it with the observations. As stated in Section \ref{ssec:weather}, we used as observation the E-OBS dataset for the years 2011-2012. 
\begin{figure}
\centering
\subfloat[Average] {
\label{fig:avgt2m}
	\includegraphics[width=2.2in]{{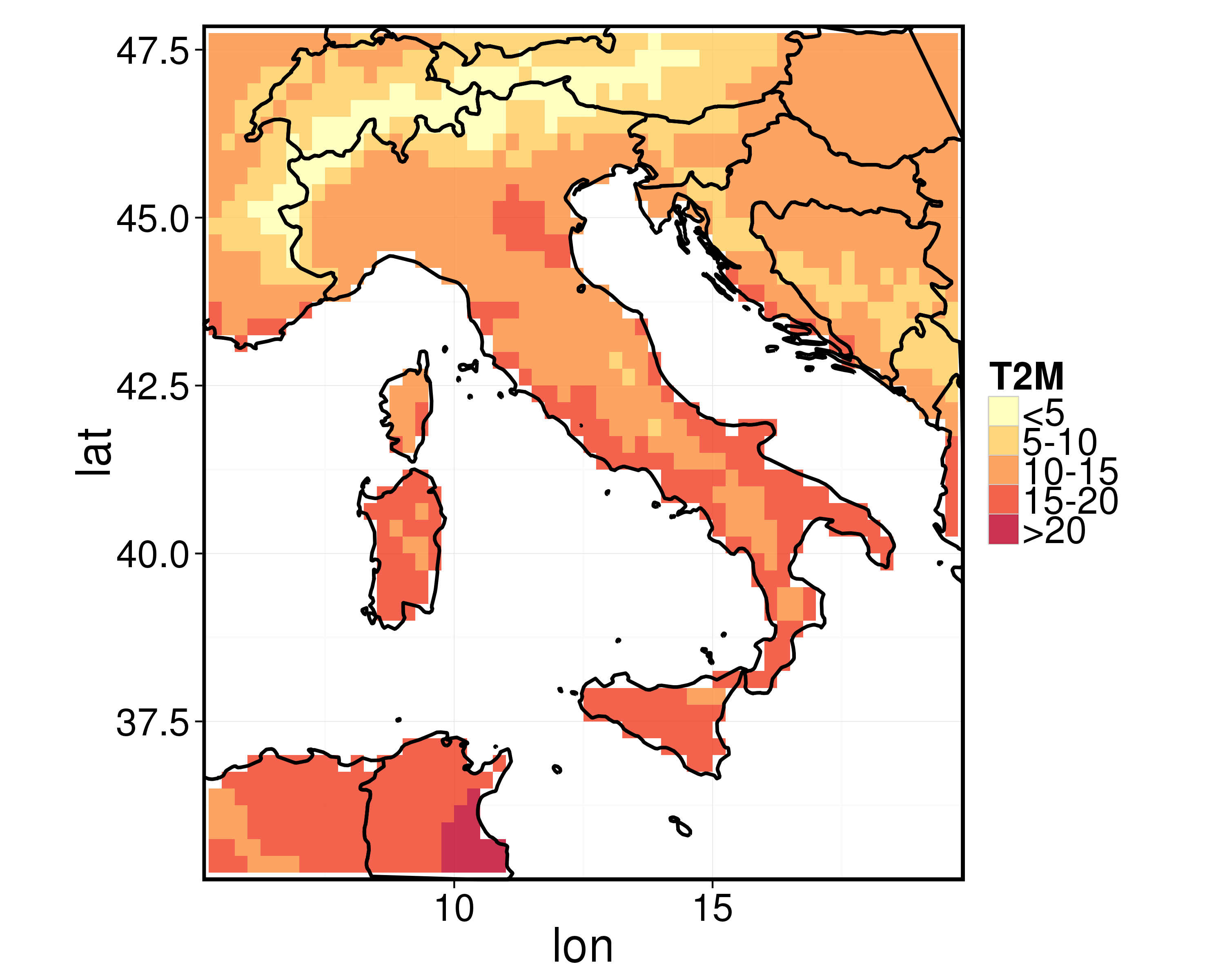}}
}
\subfloat[Coefficient of Variation] {
\label{fig:cvt2m}
	\includegraphics[width=2.2in]{{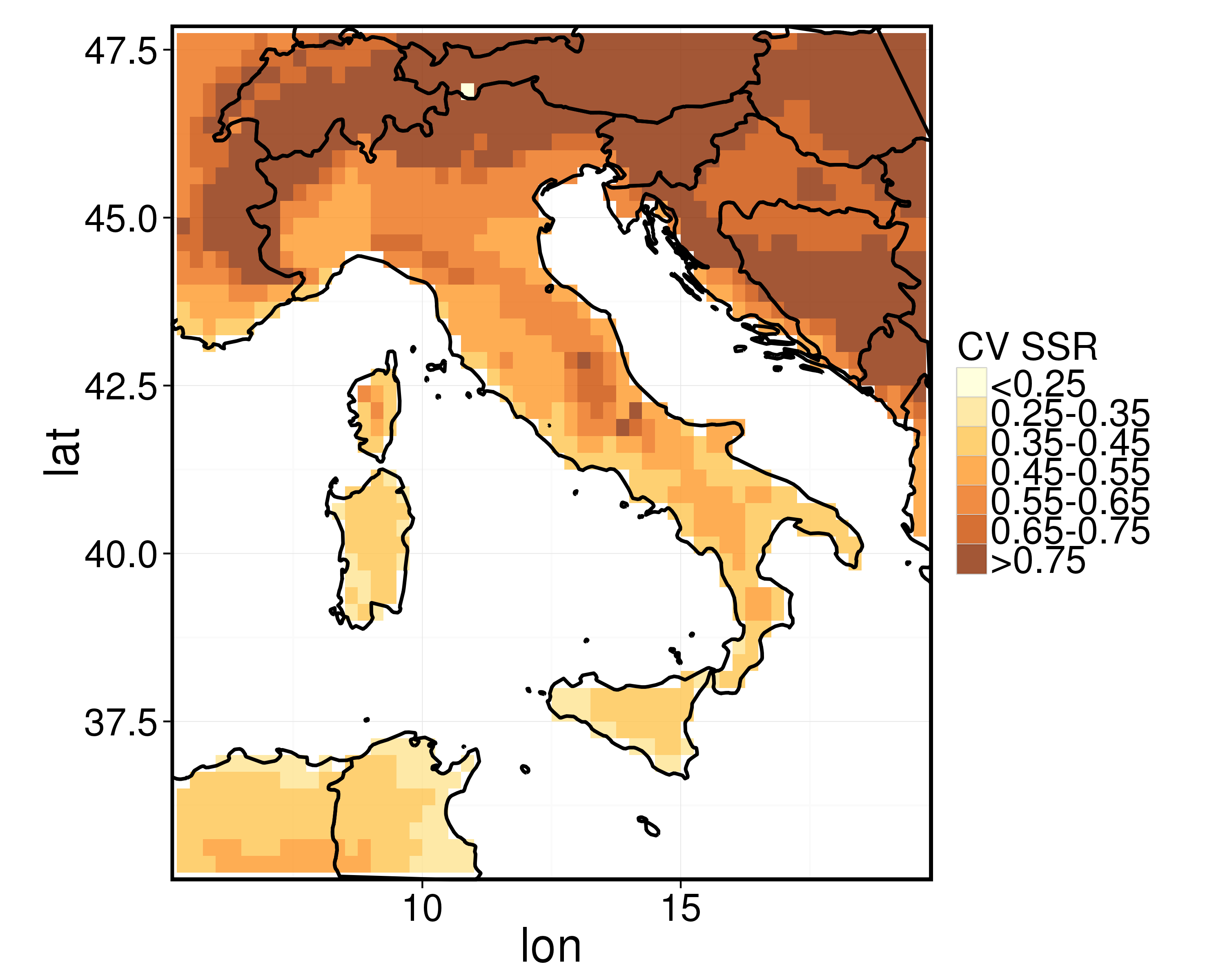}}
}
\caption{Air Temperature statistics for the years 2011-2012 from E-OBS dataset. Both the statistics highlight clearly the Italian coastal areas (higher average temperature and lower variability) and the mountain areas (Alps and Apennines with lower temperature and higher variability).}
\label{fig:comparet2m}
\end{figure}

\begin{figure}
\centering
\subfloat[NORTH] {
	\label{sfig:t2mdensitynorth}
	\includegraphics[height=1.25in]{{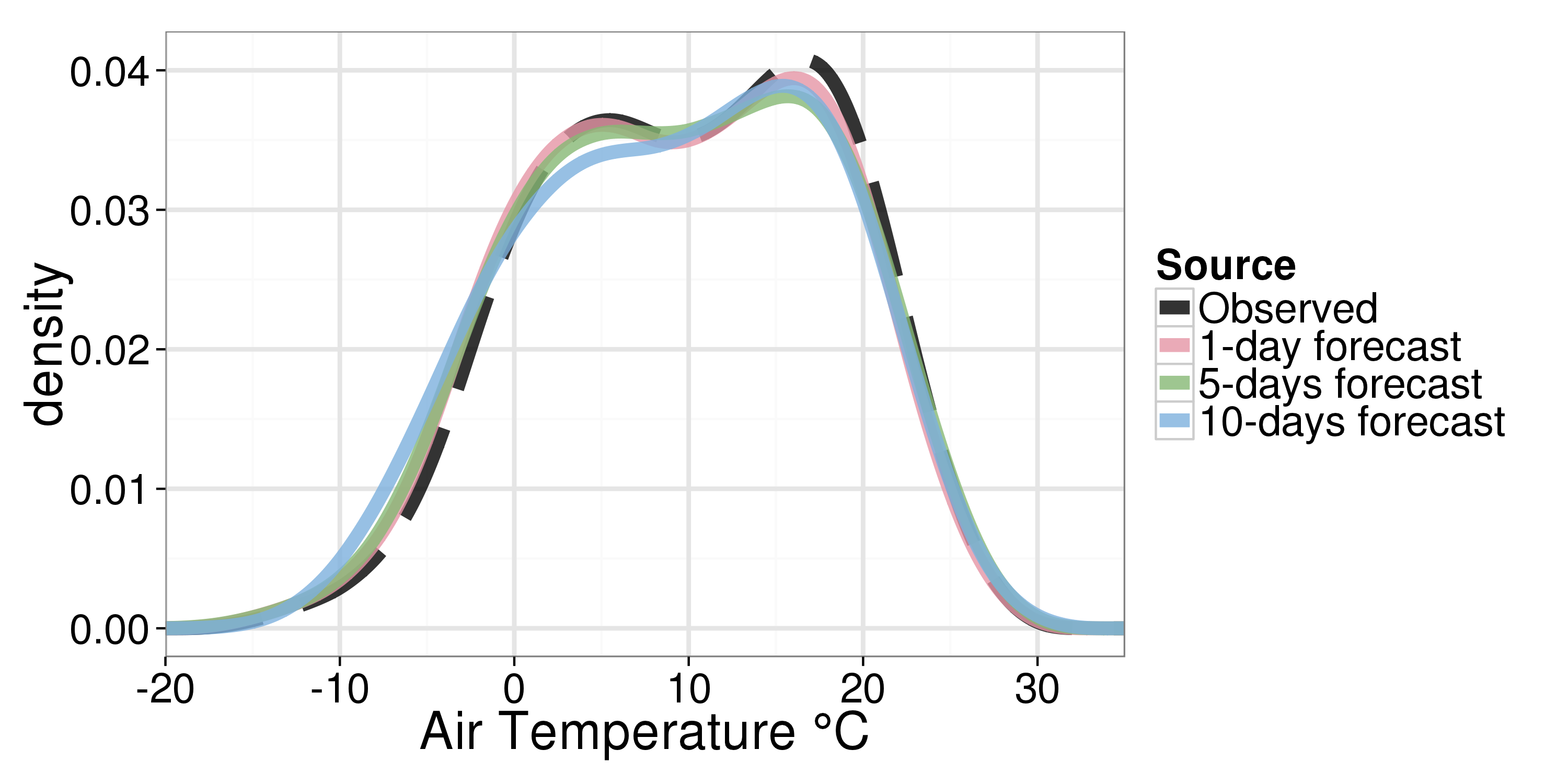}}
} \hfill
\subfloat[SOUTH] {
	\label{sfig:t2mdensitysouth}
	\includegraphics[height=1.25in]{{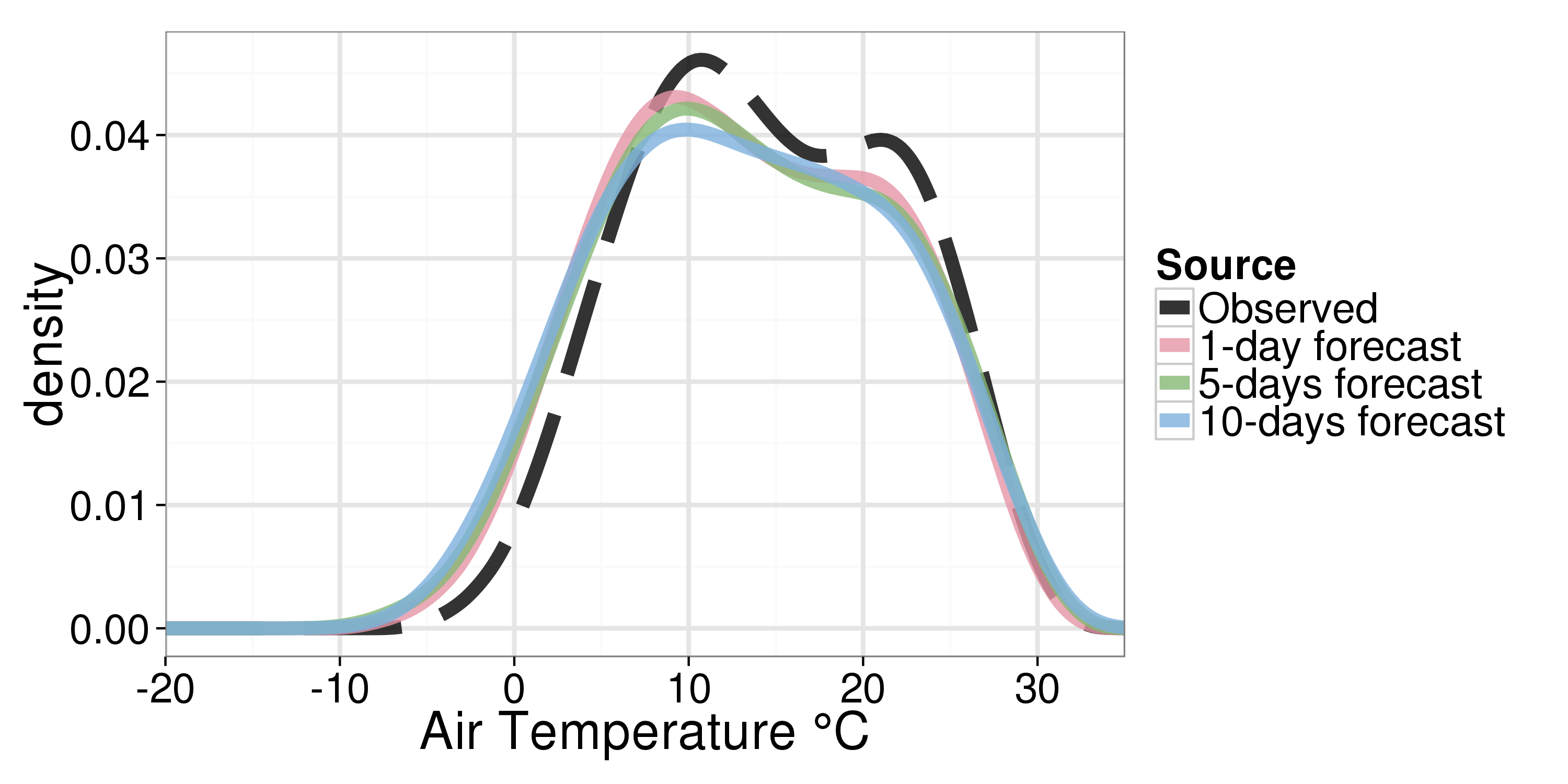}}
}
\caption{Comparison of kernel density estimation of the observed temperature with the predictions at three lead-times (one, five and ten days). }
\label{fig:densityT2M}
\end{figure}

Figure \ref{fig:comparet2m} shows descriptive statistics of observed temperature over Italy for the years 2011-2012. The coefficient of variation (Figure \ref{fig:cvt2m}) clearly follows Italian orography, with the higher variability of temperature mostly in the mountain areas. The density plot of observed and predicted temperature (Fig. \ref{fig:densityT2M}) shows a higher correspondence of forecasts with respect to the similar plot for solar radiation in Figure \ref{fig:densitySSR}. 

\section{Modelling PV production using satellite data}
\label{sec:modeling}
To perform a forecast of the solar power production we first need to find an accurate relationship between daily meteorological variables (here solar radiation and temperature) and power production. We need to find a function $f_i$ for each PV plant with the following form:
\begin{equation}
 \hat{y} = f_i(\mathrm{SSR}, \mathrm{T})
 \end{equation} 
with $\hat{y}$ the predicted power output and SSR and T respectively the surface solar radiation and the ambient temperature available for the i-th PV plant. This function aims to model the relationship between the weather variables and the electricity produced, trying to minimise the error between observed and estimated values. A black-box approach will focus at the same time on the minimization of the modelling error and on the maximization of the generalization, i.e. the capability of giving consistent outputs with unseen inputs. 
Given the absence of on-site measurements, here we consider as inputs the bilinear interpolation among the four nearest grid points of solar radiation and temperature data. 

Although the photovoltaic process is non-linear, it is a good practice to start with the simplest model for the $f$ function, a linear regression model with the following form:
\begin{equation}
\hat{y} = a_1 \mathrm{SSR} + a_2 \mathrm{T} + a_3
\end{equation}

Minimizing the error through Ordinary Least Squares, we obtain an average MdAPE of 12.4\% on cross-validation. A k-fold (with $k = 10$) cross-validation procedure here is used: as first step we divide the available dataset in $k$ subsamples of equal size, and then for $k$ times the chosen model is calibrated using $k-1$ subsets and then tested on the remaining one. At the end of the $k$ steps, the cross-validation error is given as the average of all the $k$ obtained errors.

Afterwards, we use a non-linear model, a Support Vector Machine (SVM). 

SVMs were developed by Cortes \& Vapnik \cite{cortes95,vapnik00} for binary classification and then extended to regression problems (Support Vector Regression). The idea behind the support vector-based methods is to use a non-linear mapping $\Phi$ to project the data into a higher dimensional space where solving the classification/regression task is easier than in the original space. 

In our case, we used a Support Vector Regression method called $\epsilon$-SVR \cite{drucker97}, which tries to find a function $f(x) = \langle w, \Phi(x) \rangle + b$ that has at most $\epsilon$ deviation from the target values. A $\epsilon$-SVR model has three parameters: the regularization parameter $C$, the $\epsilon$ value, and the width of the radial kernel $\gamma$. 

For each PV plant we chose the optimal parameters of the SVR model applying a grid search among 75 combinations of $C \in [10^{-2}, 10^2]$, $\epsilon \in [10^{-2}, 1]$) and $\gamma \in [2^{-2}, 2^2]$. After the parameters' selection, as for the linear models, we compute the cross-validation error. We obtain an average MdAPE of 7.6\%, about the 40\% lower than in the linear case. This improvement is expected, given the highest modelling power due to the inherent non-linearity of SVR with respect to linear regression.

\begin{figure}
\centering
	\includegraphics[width=9cm]{{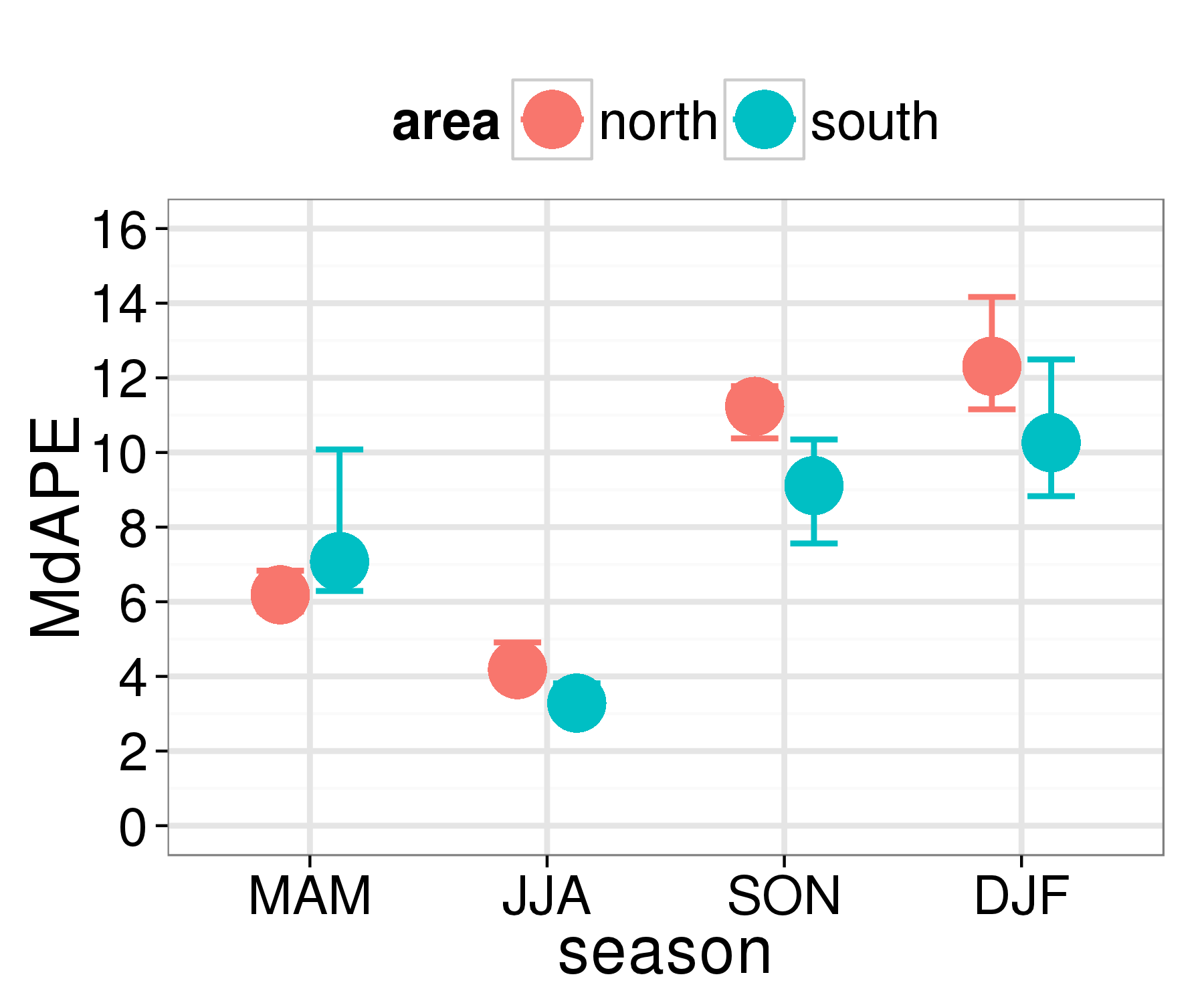}}
\caption{Cross-validation modelling error for SVM using observed meteorological variables (satellite solar~radiation and EOBS temperature). Error bars represents the interquartile range (IQR). The model is able to model the power production better in the South Italy than in the North, due to the lower weather variability, except during Spring.}
\label{fig:seasonsvm}
\end{figure}

Aggregating the PV plants by North and South (see Sec. \ref{ssec:proddata}) we come to the modelling errors shown in Figure \ref{fig:seasonsvm}. We observe how the percentage error is lowest during summer for entire Italy, and, except for Spring, we get for South Italy lower errors among all the seasons. 

\section{Short-term forecast of solar power production}
\label{sec:forecast}
In this section we assess the forecasting skill using the SVM models created in the previous section driven by predicted weather variables instead of observations. 

As summarised in Table \ref{tab:datasets} and explained in Section \ref{ssec:weather}, for the prediction we use the meteorological data coming from the ECMWF operational forecasts. As for the modelling part, for each PV plant we use bilinear interpolation of the nearest four grid points as input variables. 

\begin{figure}
\centering
\subfloat[MdAPE percentage error] {
\label{sfig:mdapeprediction}
	\includegraphics[height=2in]{{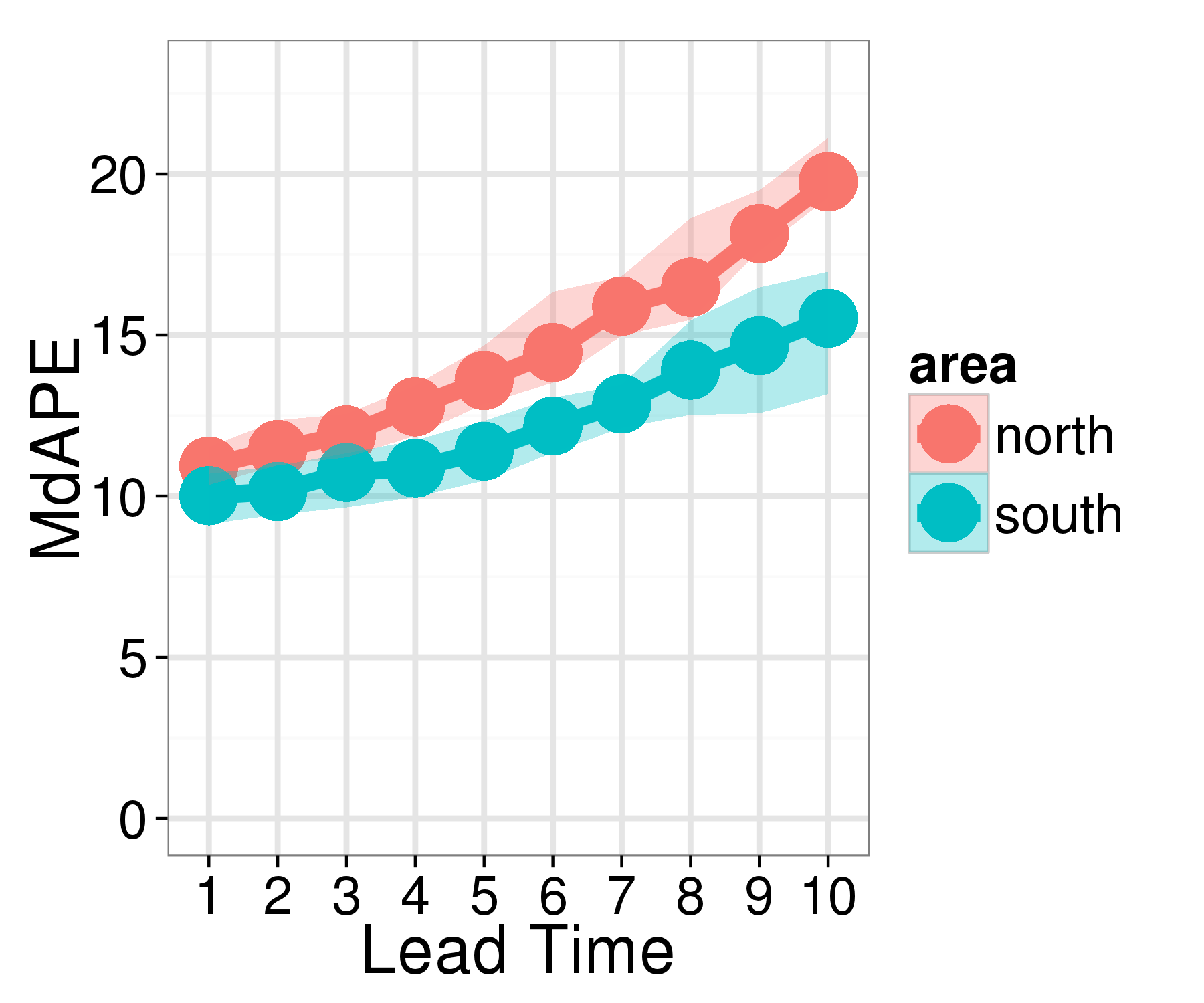}}
}
\subfloat[Correlation] {
\label{sfig:correlationprediction}
	\includegraphics[height=2in]{{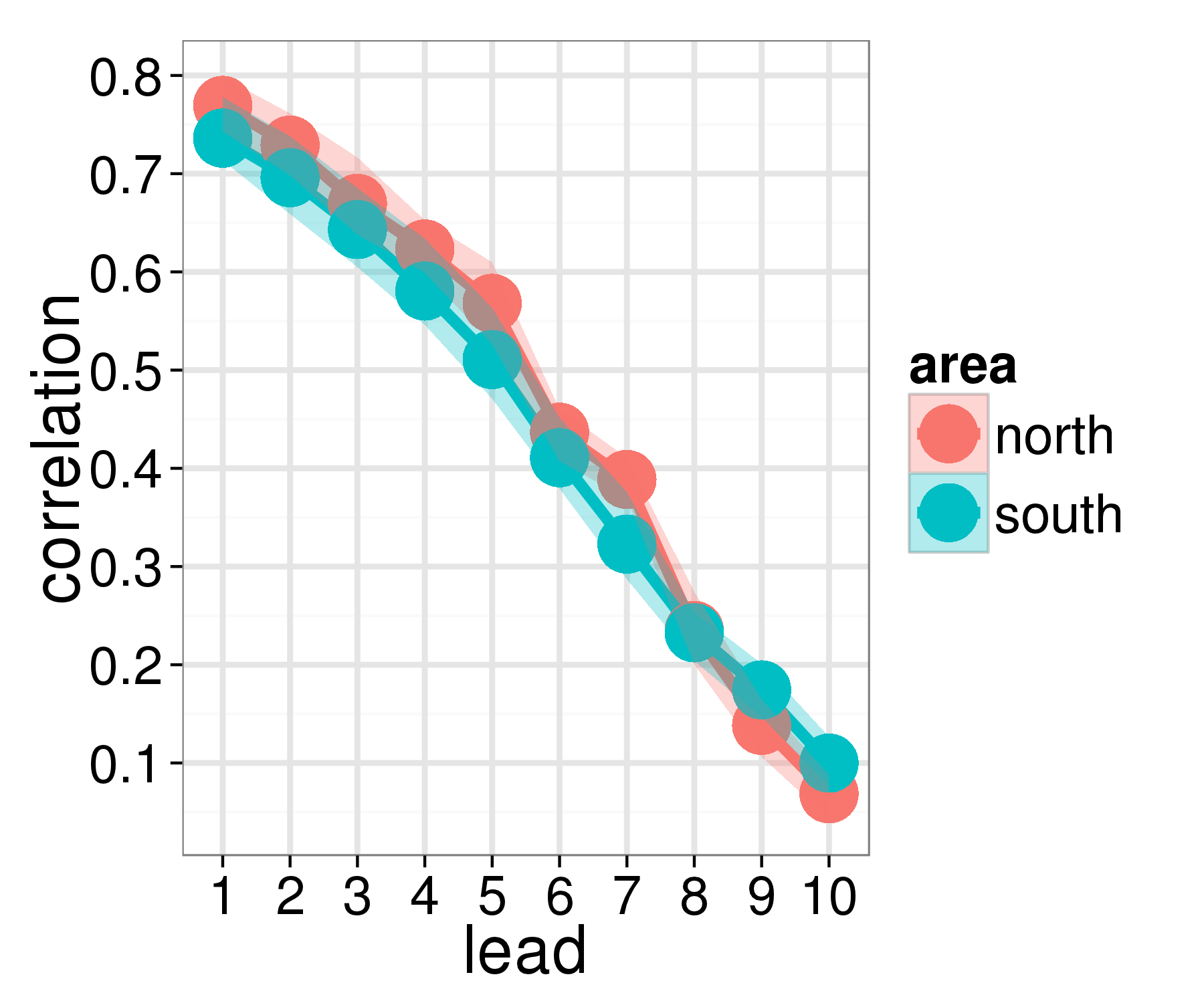}}
	}
\caption{PV power production forecast for SVM using predicted meteorological data. Shaded area represents interquartile range (IQR). }
\label{fig:leadtimesvm}
\end{figure}

For each day of lead time we show the error of the power production in Figure \ref{sfig:mdapeprediction} while the correlation between predicted and observed output is shown in Figure \ref{sfig:correlationprediction}. The minimum error is with one day of lead time ($10-12\%$) and it grows steadily up to $15-20\%$ with ten days of lead time. In all the cases the prediction of the PV plants in the South of Italy is more accurate than in the North, and we observe that the interquartile range also increases with the lead time, evidencing the higher uncertainty due to the weather forecasts at bigger lead times. 
Looking at the correlation we can see that with one day of lead time for both the cases it is in the range $0.7-0.8$ while at ten days it drastically decreases below $0.1$. 

\begin{figure}
\centering
\includegraphics[width=11.5cm]{{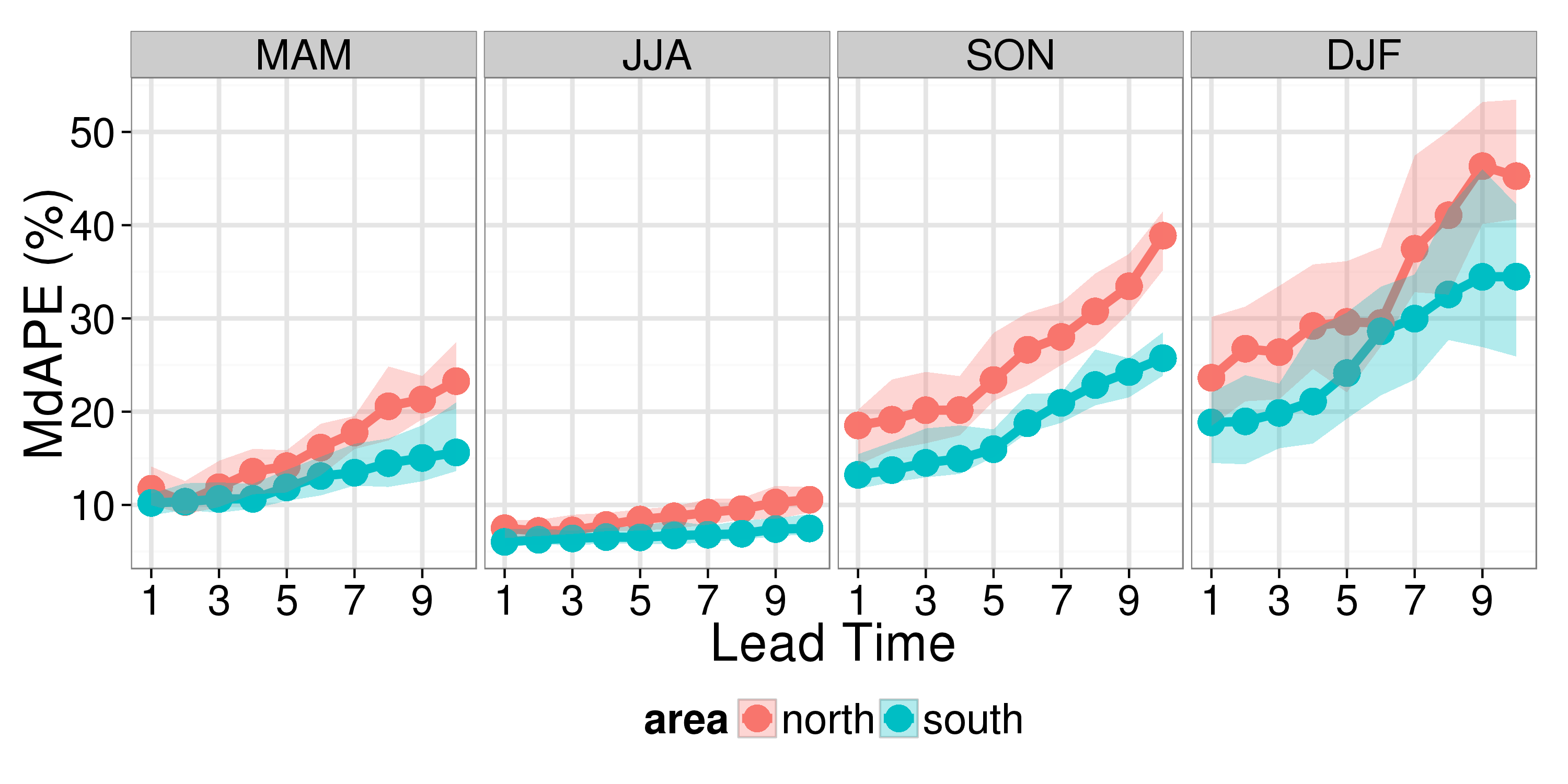}}
\caption{Prediction error (median percentage absolute error) for SVM using forecasted weather data by season. Shaded area represents interquartile range (IQR). The evident error differences among the seasons (especially between summer and the other season) is due to the weather variability and then to the capability of the weather forecast models to predict effectively the meteorological predictors used as inputs for the SVM. }
\label{fig:leadtimesvmbyseason}
\end{figure}

The error analysis can be improved grouping the errors by season, as in Figure \ref{fig:leadtimesvmbyseason}. In this figure is clearly evident the difference of errors between spring /summer, where it is common to have clear sky in most of the country, and autumn/winter, where the errors reach about the $50\%$ of MdAPE. 

\begin{figure}
\centering
\subfloat[NORTH] {
	\includegraphics[height=1.75in]{{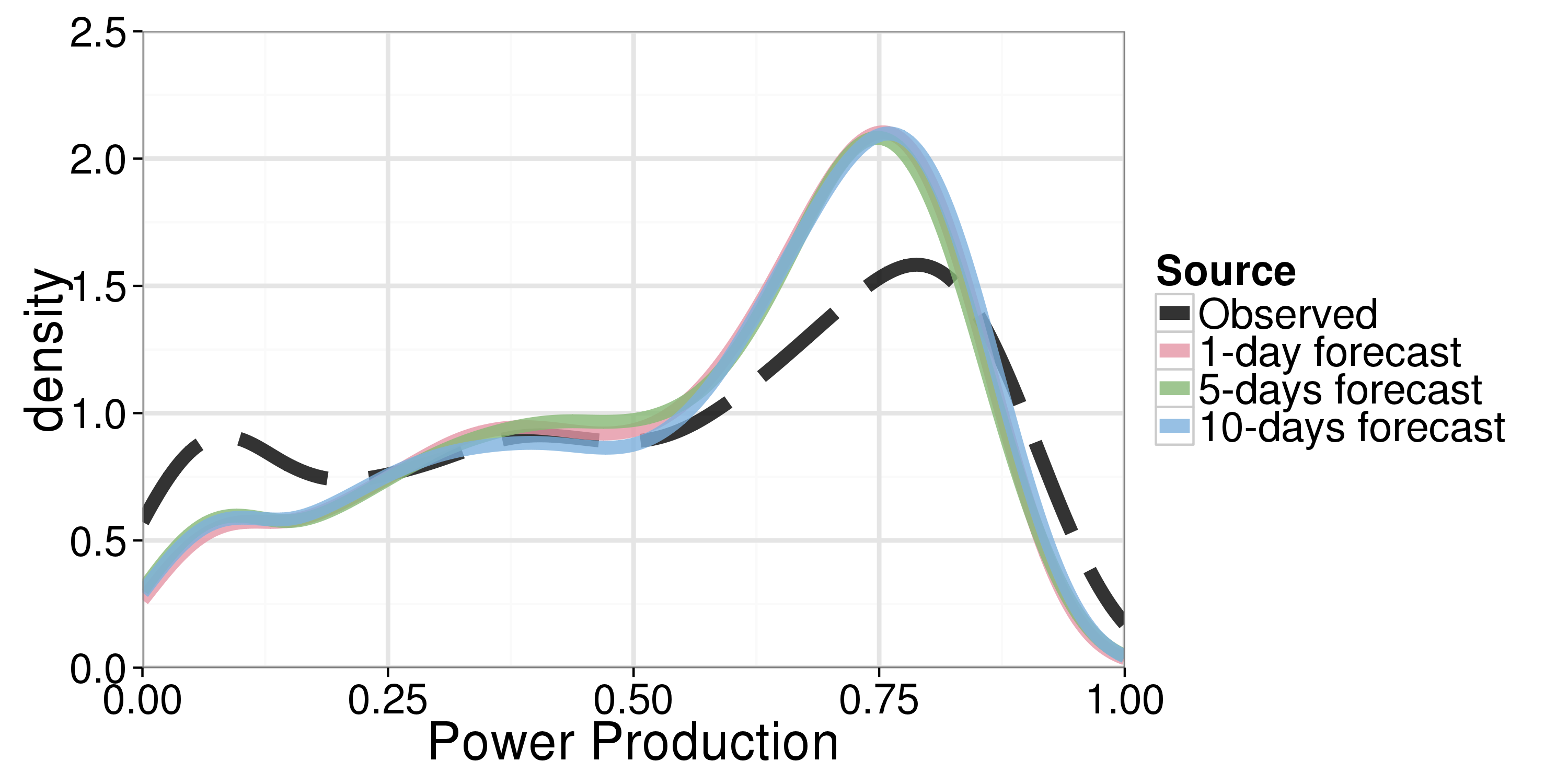}}
} \hfill
\subfloat[SOUTH] {
	\includegraphics[height=1.75in]{{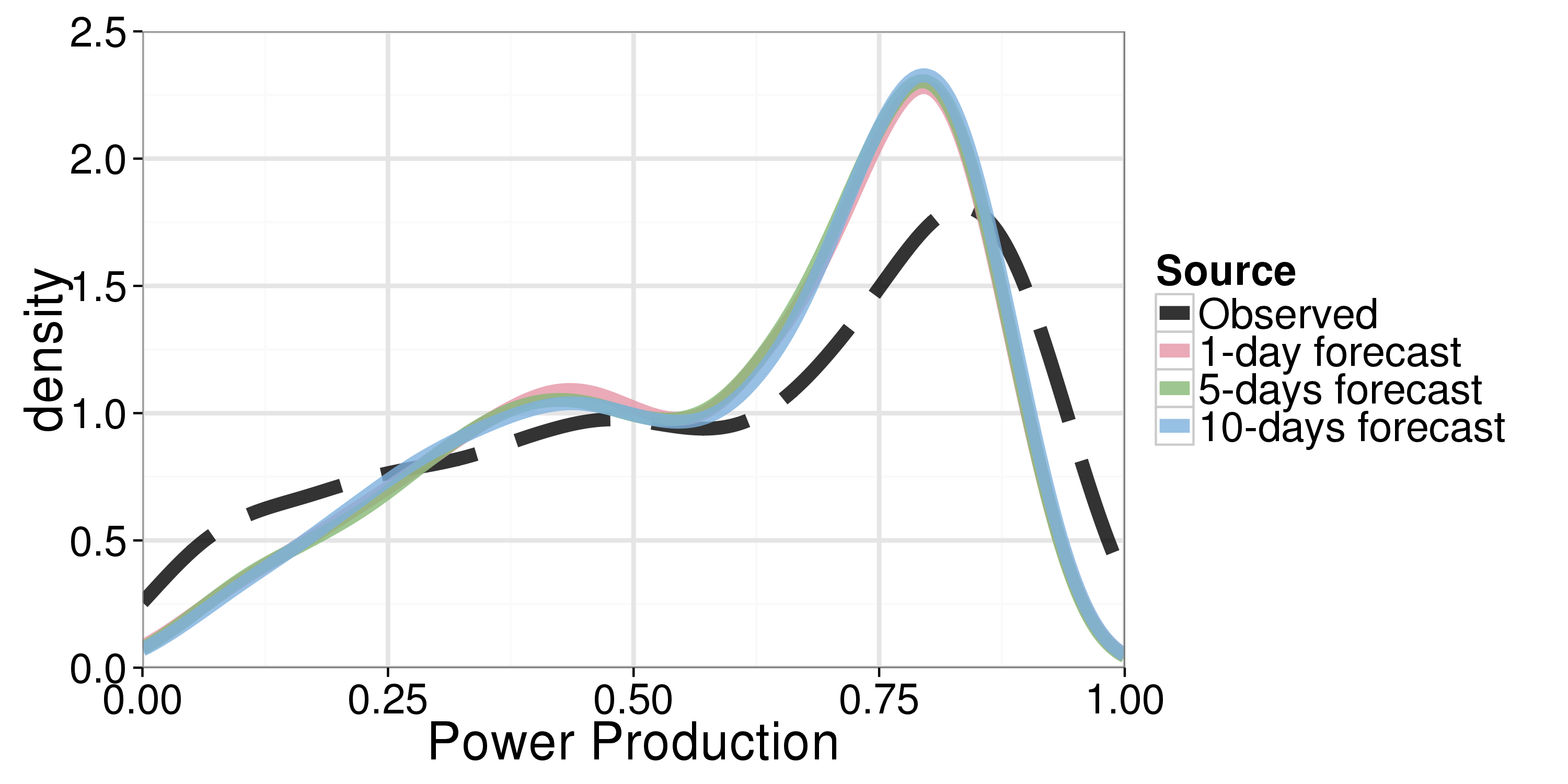}}
}
\caption{Comparison of kernel density estimation of the normalized solar power production with the predictions at three lead-times (one, five and ten days). The SVM model tends to underestimate the power production in both the geographical domains and basically the power distributions of the three lead-times are hardly distinguishable.}
\label{fig:densityPROD}
\end{figure}

Finally, the plot shown in Figure \ref{fig:densityPROD} makes evident how in both the cases the prediction densities of the three lead times provided look very similar, evidencing a general tendency to underestimate high yields.

\section{Conclusions}
\label{sec:conclusions}

In this paper, we have shown an assessment about the short-term predictability of photovoltaic daily power production over Italy without the use of on-site measurements. A detailed analysis of the weather forecast performances of solar radiation and temperature has been performed, in order to get a deeper understanding of the solar PV forecast performances.

Using a Support Vector Machine model, we have analysed the modelling error of power production using solar radiation and temperature observations, respectively from satellite and weather stations. We have compared the prediction error obtained using weather forecasts as inputs for lead times between one and ten days. 

The results can be outlined as follows:
\begin{enumerate}
\item Without using on-site measurements and using instead meteorological information provided by satellite and weather stations interpolated on the PV plant location, we obtain an average cross-validation percentage error (MdAPE) of 12.4\% using a linear model and 7.6\% a SVM.
\item Solar power production modelling on Italy was found to be more accurate during summer than in the rest of the year: the error is below the 5\% when we use observed meteorological data as predictors and below the 12\% for the entire prediction range when we use forecasted predictors.
\item The prediction results for the PV plants in the South Italy were comfortably superior than those in the North, mainly due to the lower weather variability in the southern part of the country. 
\end{enumerate}

\begin{figure}
\centering
\includegraphics[width=11.5cm]{{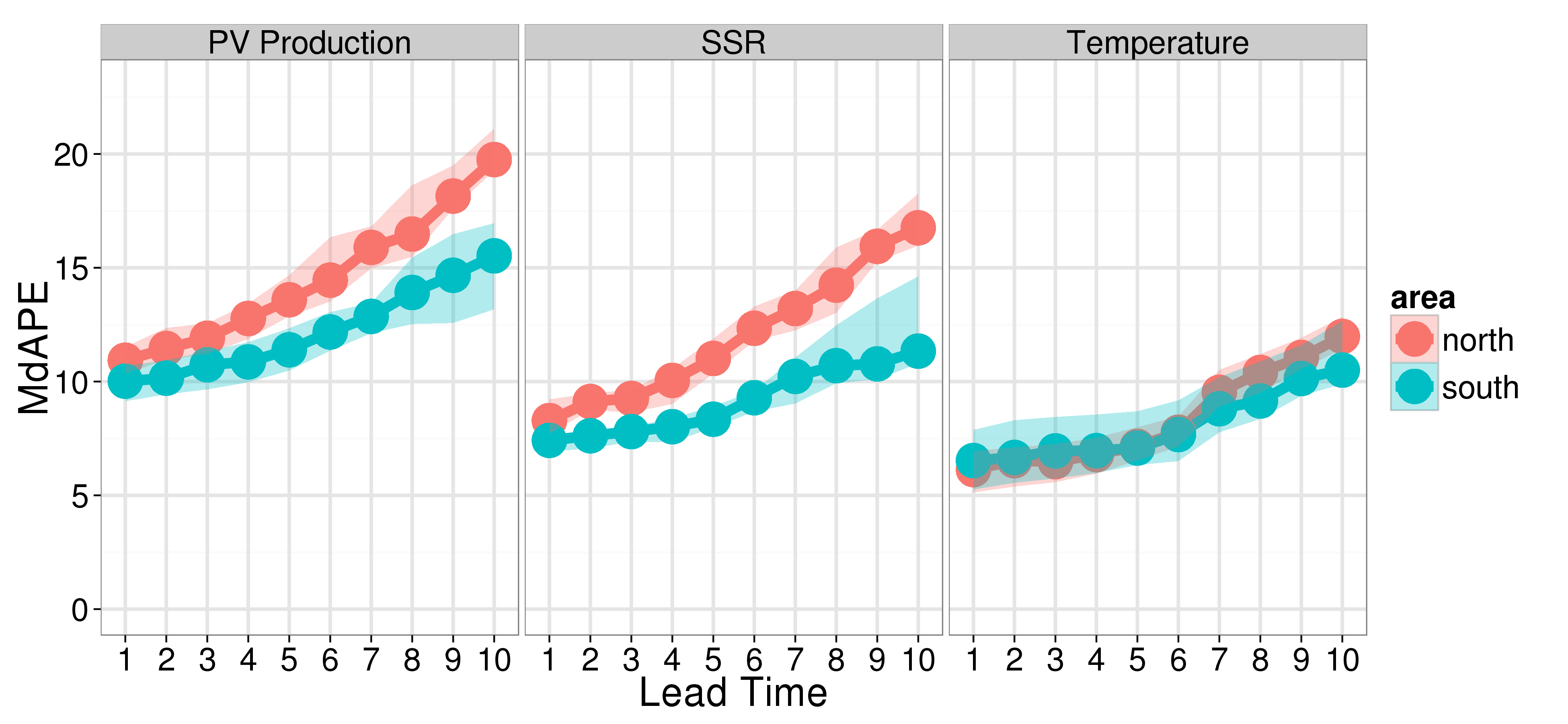}}
\caption{Prediction error (median percentage absolute error) for SVM using predicted meteorological data compared with the prediction errors of the inputs (solar radiation and temperature). Shaded area represents interquartile range (IQR). }
\label{fig:svmerrorfacet}
\end{figure}

Uncertainty due to the absence of information related to local phenomena (e.g. orography, shading effects) becomes certainly critical in predicting PV power production, especially for the higher lead times. The uncertainty due to weather forecasts can be estimated observing the ``distance'' between the modelling (Fig. \ref{fig:seasonsvm}) and prediction (Fig. \ref{fig:leadtimesvm}, \ref{fig:leadtimesvmbyseason}) errors. The former in fact represents the error due to model limitations and observation errors due to interpolation. When we apply the same model for the forecasting, we add then the error due to the weather predictions, the same error discussed in Sections \ref{sec:solarpred} and \ref{sec:temppred}. Figure \ref{fig:svmerrorfacet} tries to represent this ``uncertainty propagation'' showing the relationship between the PV production error (the same as in Fig. \ref{sfig:mdapeprediction}) and the forecast error of the used meteorological predictors (solar radiation and temperature). 

These results demonstrate the potentiality in using black-box approach in spite of the absence of on-site measurements. 
 
\section{Acknowledgments}
We thank TERNA for providing photovoltaic data. EUMETSAT Satellite Application Facility on Climate Monitoring (CM SAF) intermediate products were used by permission of Deutscher Wetterdiens. We also acknowledge the E-OBS dataset from the EU-FP6 project ENSEMBLES (http://ensembles-eu.metoffice.com) and the data providers in the ECA\&D project (http://www.ecad.eu). 

\bibliographystyle{plain}
\bibliography{pvterna}

\end{document}